\documentclass{IEEEtran}
\usepackage{graphicx} 
\usepackage{booktabs}
\usepackage[table,xcdraw]{xcolor}
\usepackage{url}
\usepackage{tabularx}
\usepackage{adjustbox}
\usepackage{float}
\usepackage{amsmath}
\usepackage[caption=false,font=normalsize,labelfont=sf,textfont=sf]{subfig}
\usepackage{cite}
\usepackage{algorithm}
\usepackage{algpseudocode}
\usepackage{multirow}
\usepackage{bm}
\usepackage{hyperref}

\begin{document}

\title{AI-Powered Intracranial Hemorrhage Detection: A Co-Scale Convolutional Attention Model with Uncertainty-Based Fuzzy Integral Operator and Feature Screening}

\author{Mehdi Hosseini Chagahi, Niloufar Delfan, Behzad Moshiri, Md. Jalil Piran, Jaber Hatam Parikhan


\thanks{M. H. Chagahi and N. Delfan are with the School of Electrical and Computer Engineering, College of Engineering, University of Tehran, Tehran, Iran, (e-mail: mehdichagahi@gmail.com;
niloufardelfan@gmail.com )}

\thanks{B. Moshiri is with the School of Electrical and Computer Engineering, College of Engineering, University of Tehran, Tehran, Iran and the Department of Electrical and Computer Engineering University of Waterloo,
Waterloo, Canada, (e-mail: moshiri@ut.ac.ir)}

\thanks{M. J. Piran is with the Department of Computer Science and Engineering, Sejong University, Seoul 05006, South Korea, (e-mail: piran@sejong.ac.kr)}

\thanks{J. H. Parikhan is with the Department of Neurosurgery, Iran University of Medical Sciences}}

\maketitle

\begin{abstract}
Intracranial hemorrhage (ICH) refers to the leakage or accumulation of blood within the skull, which occurs due to the rupture of blood vessels in or around the brain. If this condition is not diagnosed in a timely manner and appropriately treated, it can lead to serious complications such as decreased consciousness, permanent neurological disabilities, or even death.The primary aim of this study is to detect the occurrence or non-occurrence of ICH, followed by determining the type of subdural hemorrhage (SDH). These tasks are framed as two separate binary classification problems. By adding two layers to the co-scale convolutional attention (CCA) classifier architecture, we introduce a novel approach for ICH detection. In the first layer, after extracting features from different slices of computed tomography (CT) scan images, we combine these features and select the 50 components that capture the highest variance in the data, considering them as informative features. We then assess the discriminative power of these features using the bootstrap forest algorithm, discarding those that lack sufficient discriminative ability between different classes. This algorithm explicitly determines the contribution of each feature to the final prediction, assisting us in developing an explainable AI model. The features feed into a boosting neural network as a latent feature space.
In the second layer, we introduce a novel uncertainty-based fuzzy integral operator to fuse information from different CT scan slices. This operator, by accounting for the dependencies between consecutive slices, significantly improves detection accuracy.
\end{abstract}


\section{Introduction}
ICH, recognized as one of the most critical and life-threatening medical conditions, can severely impact patient health and require prompt and accurate diagnosis in order to ensure appropriate treatment \cite{chen2024efficient,ferdi2024quadratic}. These types of hemorrhages are typically caused by traumatic injuries, such as head trauma, or specific medical conditions like aneurysm rupture, high blood pressure, or blood disorders \cite{nizarudeen2024comparative,malik2023computational}. This emergency situation can rapidly increase intracranial pressure and cause damage to brain tissues \cite{umapathy2023automated,angkurawaranon2023comparison}.

In the past, diagnosing such hemorrhages primarily relied on conventional imaging techniques, such as CT scans and magnetic resonance imaging (MRI). These images were reviewed by radiologists and neurologists to determine the type and severity of the hemorrhage \cite{lyu2023machine,soheili2024toward}. However, due to the reliance on the expertise and experience of physicians, these methods have always faced challenges related to accuracy and speed \cite{gudadhe2023classification,champawat2023literature}.

In the past decade, with significant advancements in artificial intelligence and machine learning (ML), research has shifted towards the development of automated models for the detection and classification of intracranial hemorrhages \cite{asif2023intracranial,yeo2023evaluation,xiao2024multi,dashtaki2024lag}. Deep learning (DL) models, such as convolutional neural networks (CNNs), pre-trained networks, and attention-based architectures, have garnered considerable attention from researchers due to their high capability in analyzing medical images \cite{chagahi2024cardiovascular,hu2024deep,delfan2024ai,delfan2024hybrid,lee2019explainable,chagahi2024enhancing}. These models can automatically analyze CT scan images and accurately detect ICH.

The study \cite{castro2024hyperbolic} presented a novel method for Multiple Instance Learning (MIL) for medical image analysis, particularly for detecting ICH in CT scans. The approach built on Gaussian Process-based MIL (VGPMIL) but introduced a new formulation using Pólya-Gamma random variables to handle the logistic function more efficiently. The reformulation, called PG-VGPMIL, was mathematically equivalent to the original VGPMIL model but provided better computational tractability.

Perez et al. \cite{perez2024end}, presented a novel end-to-end model combining CNN, attention mechanisms, and gaussian processes for MIL to detect ICH from CT scans. The method overcame challenges in the manual annotation of large datasets by using scan-level labels instead of slice-level labels. The CNN was employed to extract features from each CT slice, which were then processed by an attention mechanism that assigned weights to highlight important slices. These weighted feature vectors were passed into a sparse gaussian process for probabilistic classification.

The work \cite{d2024accuracy} evaluated a novel DL algorithm based on the Dense-UNet architecture for detecting ICH in non-contrast CT (NCCT) head scans after traumatic brain injury. The algorithm processed CT scans by segmenting the brain using anatomical landmarks and performed volumetric segmentation to detect hemorrhage. The algorithm used a 3D Dense-UNet architecture, which introduced densely connected layers to enhance segmentation quality by allowing each layer to receive inputs from all previous layers and pass outputs to subsequent layers.

Sindhura et al.  \cite{sindhura2024fully} proposed a fully automated sinogram-based DL model for the classification of ICH directly from sinogram data, eliminating the need for the time-consuming reconstruction phase of CT scans. The method involved a two-stage approach: the first stage used a U-Net-based model to synthesize sinograms equivalent to intensity-windowed CT scans that are commonly used in brain imaging. The second stage was a detection module composed of a CNN and a recurrent neural network (RNN), where the CNN extracted features from the sinograms, and the RNN (Bi-GRU) captured spatial information across neighboring slices to improve accuracy in detecting hemorrhages.

Shah and Jaimin \cite{shah2024human} evaluated the usability and human factor engineering of a ML-powered Near-Infrared Spectroscopy (mNIRS) device, designed for point-of-care detection of ICH in patients with traumatic brain injury. The mNIRS device leveraged near-infrared light to non-invasively assess brain tissues, detecting extravascular bleeds by measuring light absorption and scattering through the scalp and skull.

The study \cite{lin2024semi}, introduced a semi-supervised learning model for the detection and segmentation of ICH from head CT scans. The proposed method combined labeled and unlabeled data to improve model generalizability, particularly on out-of-distribution datasets. The semi-supervised model showed significant improvements in generalizability compared to a fully supervised baseline model, achieving higher accuracy in both classification (AUC 0.939 vs. 0.907) and segmentation (Dice similarity coefficient of 0.829 vs. 0.809).

Ragab and Mahmoud \cite{ragab2023political} introduced the political optimizer (PO) with deep learning for ICH Diagnosis. The method began with bilateral filtering for preprocessing CT images, which helped in removing noise while preserving the edges. For feature extraction, the Faster SqueezeNet model was used to capture relevant features from the images. Finally, a denoising autoencoder model was employed for classification, with hyperparameter optimization performed by the PO algorithm.

The work \cite{cortes2023deep} proposed a DL model for ICH detection using CT scans. The model was based on a combination of two neural network architectures: ResNet and EfficientDet. ResNet addressed challenges such as vanishing or exploding gradients, while EfficientDet focused on computational efficiency through its novel Bi-directional feature pyramid network and compound scaling method. Together, these architectures formed the backbone of the model, which optimized detection accuracy while minimizing computational resources.
Li et al. \cite{li2023deep} introduced a novel DL-enabled microwave-induced thermoacoustic tomography (MITAT) technique, leveraging a residual attention U-Net architecture for transcranial brain hemorrhage detection. The method addressed challenges related to acoustic inhomogeneity caused by the skull during transcranial imaging. MITAT combined microwave irradiation with thermoacoustic effects to capture tissue contrast and generate high-resolution brain images.
Unlike previous studies that selectively utilized only a few slices from CT scan images to detect ICH, this study, inspired by real-world practices, integrates all informative slices from a CT scan using a uncertainty-based fuzzy integral operator. This approach enables the development of a more robust and reliable detection model. By employing feature selection and combination techniques to create a latent feature space, the proposed model focuses solely on the features with the highest contribution to the final prediction. This approach significantly reduces computational costs and facilitates the development of an interpretable model.
\begin{itemize}
    \item A large dataset comprising head CT scan images from two medical centers is collected and meticulously labeled.
    \item Data preprocessing involves the removal of heterogeneous color regions and the reduction of image dimensions to focus on the primary brain tissue using background subtraction and binary masking techniques.
    \item  Permutation technique is employed to assess the importance of each feature, allowing for the identification and removal of features that lack sufficient discriminative power.
    \item An uncertainty-based fuzzy integral operator is introduced, capable of combining multiple CT scan slices for enhanced diagnostic accuracy.
\end{itemize}
The structure of this study is: Section \ref{Methodology} summarizes the dataset, highlighting image preprocessing. It then discusses the processes of feature fusion and feature importance, concluding with an in-depth analysis of the proposed system’s architecture. Section \ref{Results} presents the findings and their analysis. Section \ref{recomm} explores potential avenues for future research. Finally, Section \ref{Conclusion} offers concluding remarks.

\section{Fuzzy-based ICH Detection Framework}
\label{Methodology}
\begin{figure}[t]
\centering
\includegraphics[scale=0.25]{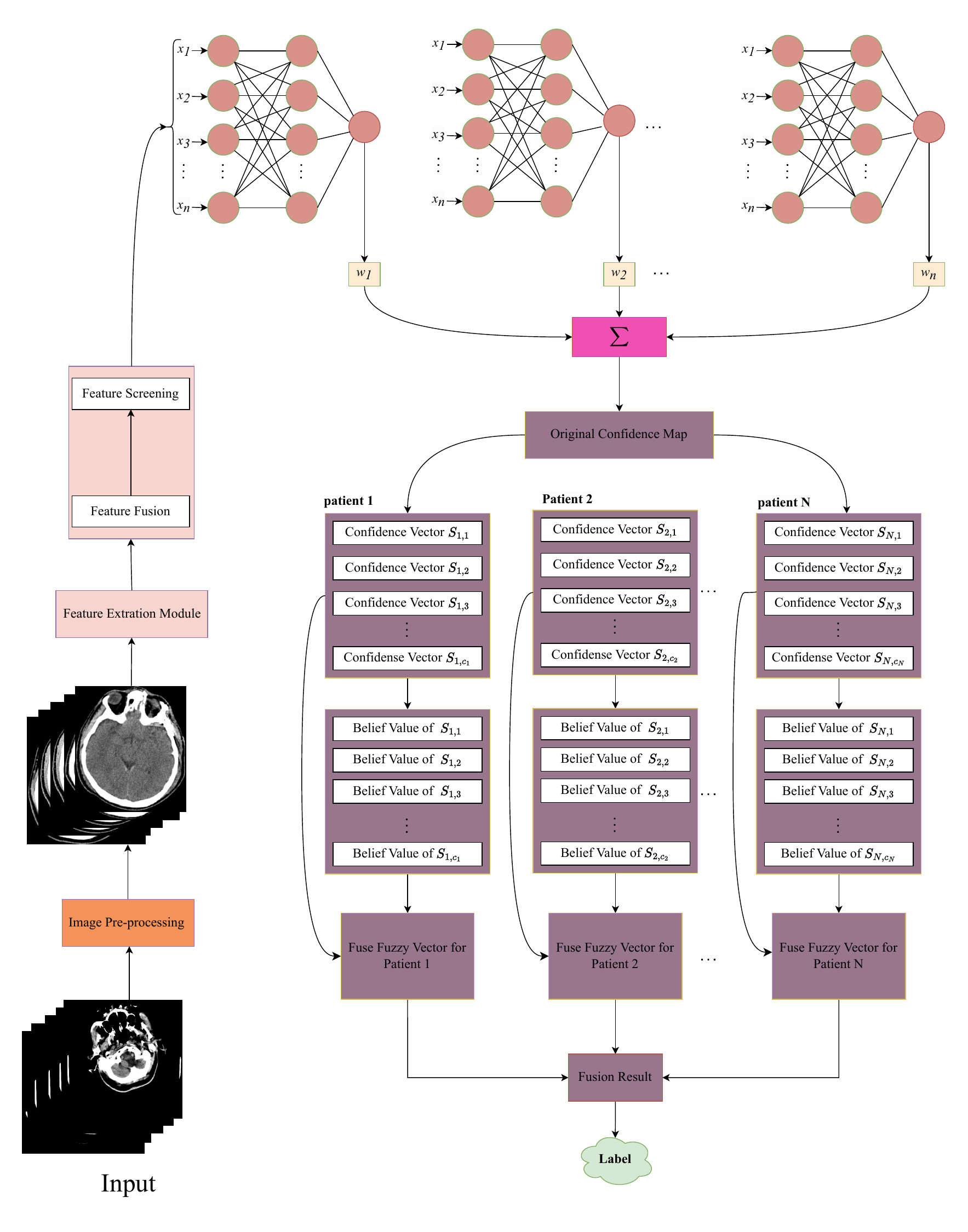}
\caption{The proposed architecture for detecting ICH.}
\label{main_arch}
\end{figure}

Fig. \ref{main_arch} illustrates the overall architecture of the proposed system for detecting ICH and SDH. Initially, the raw images undergo essential preprocessing. Following this, feature extraction is performed, and subsequently, feature fusion and selection techniques apply to ensure that only the most discriminative features are kept for input to the boosting neural network. The architecture estimates the probability of each slice belonging to specific classes. Finally, slices from the same CT scan are aggregated using a fuzzy integral operator based on uncertainty, resulting in the determination of the final label for the CT scan. In the following, we delve into the details of each component of the proposed system.
\subsection{Data acquisition and labeling}
\begin{figure*}[t]
    \centering
    \subfloat[CT scan of a normal brain without any signs of hemorrhage.]{\includegraphics[width=0.25\textwidth]{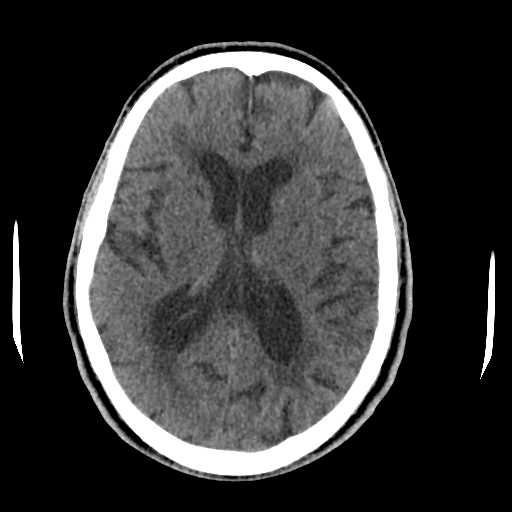}} 
    \subfloat[CT scan of a brain with ICH]{\includegraphics[width=0.25\textwidth]{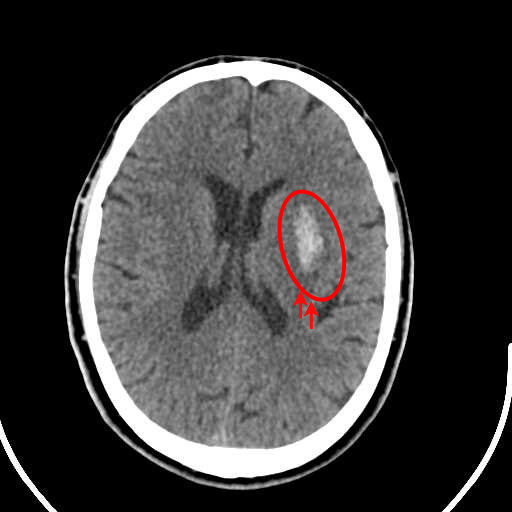}}
    \caption{(a) CT scan of a normal brain without any signs of hemorrhage, (b) CT scan of a brain with ICH, illustrating visible regions of bleeding.}
    \label{NICH}
\end{figure*}

\begin{figure*}[t]
    \centering
    \subfloat[Brain CT scan showing ASDH]{\includegraphics[width=0.25\textwidth]{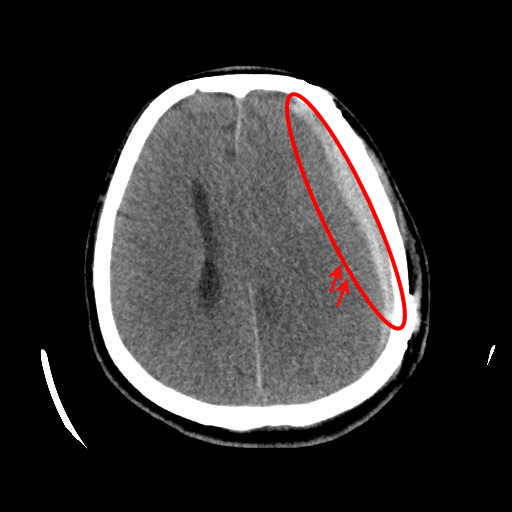}}
    \subfloat[Brain CT scan showing CSDH]{\includegraphics[width=0.25\textwidth]{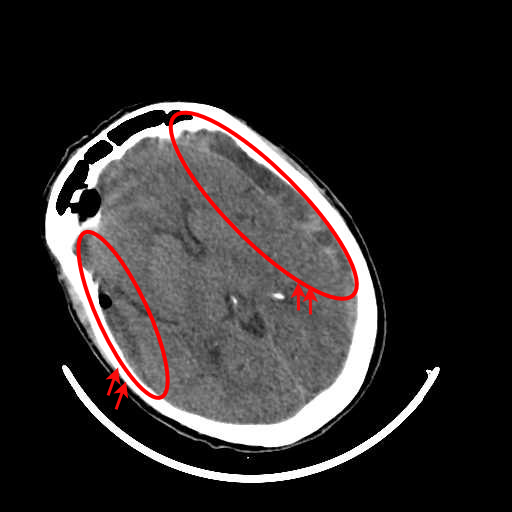}}
    \caption{(a) Brain CT scan showing ASDH, characterized by recent bleeding with high-density regions, (b) Brain CT scan showing CSDH, indicating older, low-density blood accumulation.}
    \label{SDH}
\end{figure*}

In this study, we utilize two datasets: NICH137 and SDH45. The NICH137 dataset focuses on assessing the presence or absence of ICH, while the SDH45 dataset is dedicated to identifying the type of Subdural Hemorrhage. Please see Table \ref{NumberOfSamples}  for a comprehensive overview of the two datasets. Both datasets were collected between 2018 and 2024 from two medical centers in Tehran: Rasul Akram Hospital and Firouzabadi Hospital. They were meticulously labeled by two board-certified neurosurgeons.
Labeling was initially performed by a board-certified neurosurgeon and subsequently validated by another board-certified neurosurgeon who is also a university faculty member. Fig. \ref{NICH} shows two slices from two classes: normal (no hemorrhage) and ICH (presence of at least one type of hemorrhage). Additionally, Fig. \ref{SDH} presents two slices from two classes: Acute Subdural Hemorrhage (ASDH) and Chronic Subdural Hemorrhage (CSDH). Differentiating these two types of hemorrhage is crucial, as they require distinct management and treatment approaches.

The dataset includes brain CT scans acquired using two protocols: parenchymal view and bone view. For the purpose of ICH classification, we specifically employ the parenchymal view protocol. The CT scans are got using a Siemens scanner, with each scan comprising 25 to 40 slices at a thickness of 5 millimeters.

Only the axial view for ICH classification is used due to its accessibility, as axial views are the most readily available brain CT view, whereas coronal and sagittal reconstructions are time intensive.

This research project is registered as a proposal at the Iran University of Medical Sciences and received ethical approval under code number ----. The researchers are committed to adhering to ethical principles in research and to following the Declaration of Helsinki.

\begin{table}[t]
  \centering
  \renewcommand{\arraystretch}{2}
  \caption{Summary of sample distribution across patient groups}
  \label{NumberOfSamples}
  \resizebox{0.4\textwidth}{!}{
  \begin{tabular}{|c|c|c|c|c|}
    \hline
    \multirow{2}{*}{\textbf{Sample}} & \multicolumn{2}{c|}{Binary} & \multicolumn{2}{c|}{Binary}\\
    \cline{2-5} 
    & Normal & ICH & ASDH & CSDH \\
    \hline
    \textbf{Patient} & 63 & 74 & 24 & 21 \\ 
    \hline
    \textbf{CT scan} & 200 & 200 & 55 & 48 \\
    \hline
    \textbf{Slice} & 15796 & 15800 & 1550 & 1235 \\
    \hline
  \end{tabular}
  }
\end{table}


\subsection{Image preprocessing}
The data preprocessing for preparing brain CT scan images for use in DL models is conducted in several stages. First, the head CT scan images are separated from all available images, which include various sections such as body, lungs, bones. Subsequently, these images are converted from DICOM format to the more commonly used JPEG format. At this stage, noisy slices or those lacking brain tissue are removed from the dataset to improve the quality of the input data for the models. 

To enhance the focus of DL models on the primary brain tissue and information-rich areas, background removal techniques are initially applied to identify and eliminate non-homogeneous color regions in the images. Afterward, a binary mask is employed to dynamically crop the images. The resulting cropped images contain only the main brain tissue.

\begin{figure}[t]
\centering
\includegraphics[scale=0.25]{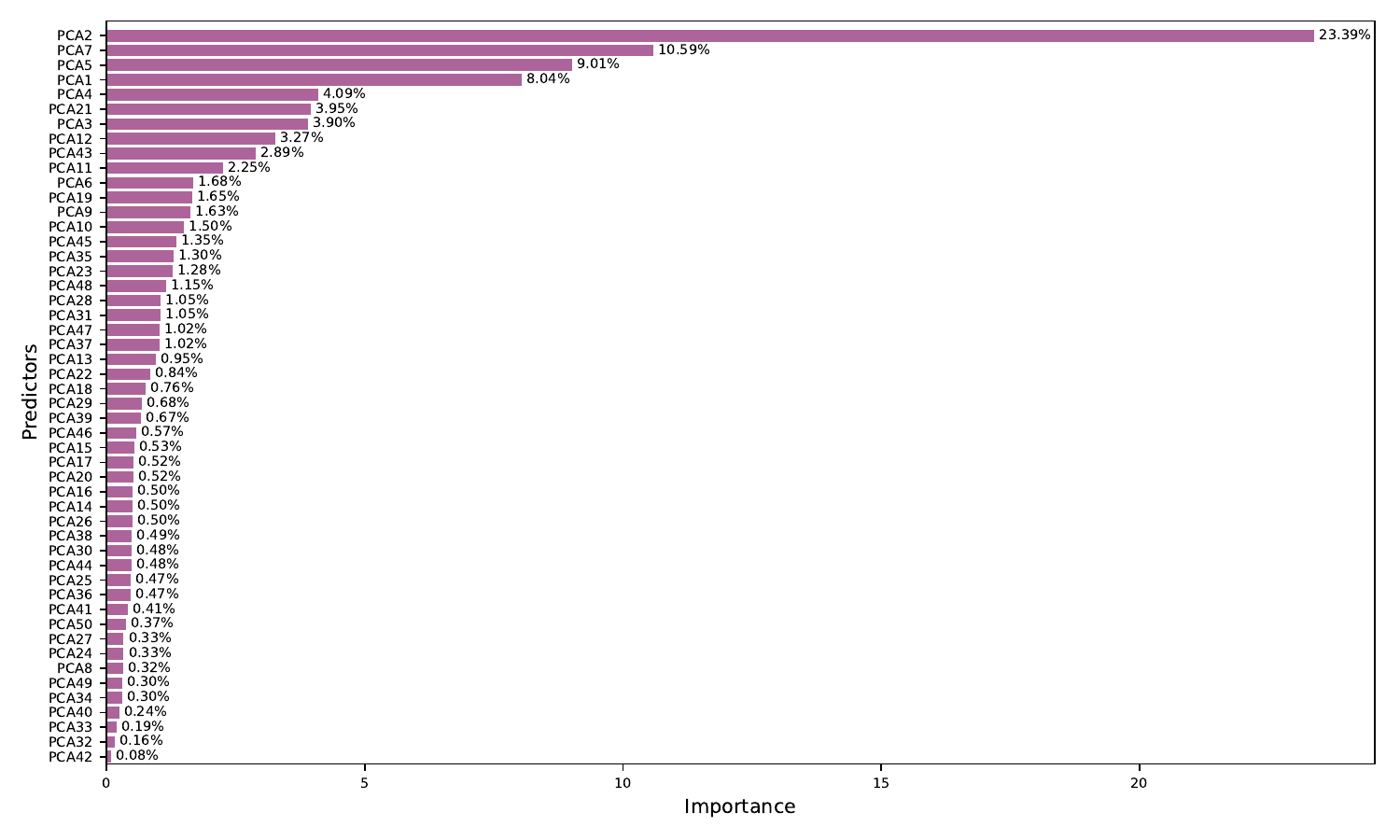}
\caption{Importance ranking of predictors based on their contribution to the ICH classification model.}
\label{NICH_featur_imp}
\end{figure}

\begin{figure}[t]
\centering
\includegraphics[scale=0.25]{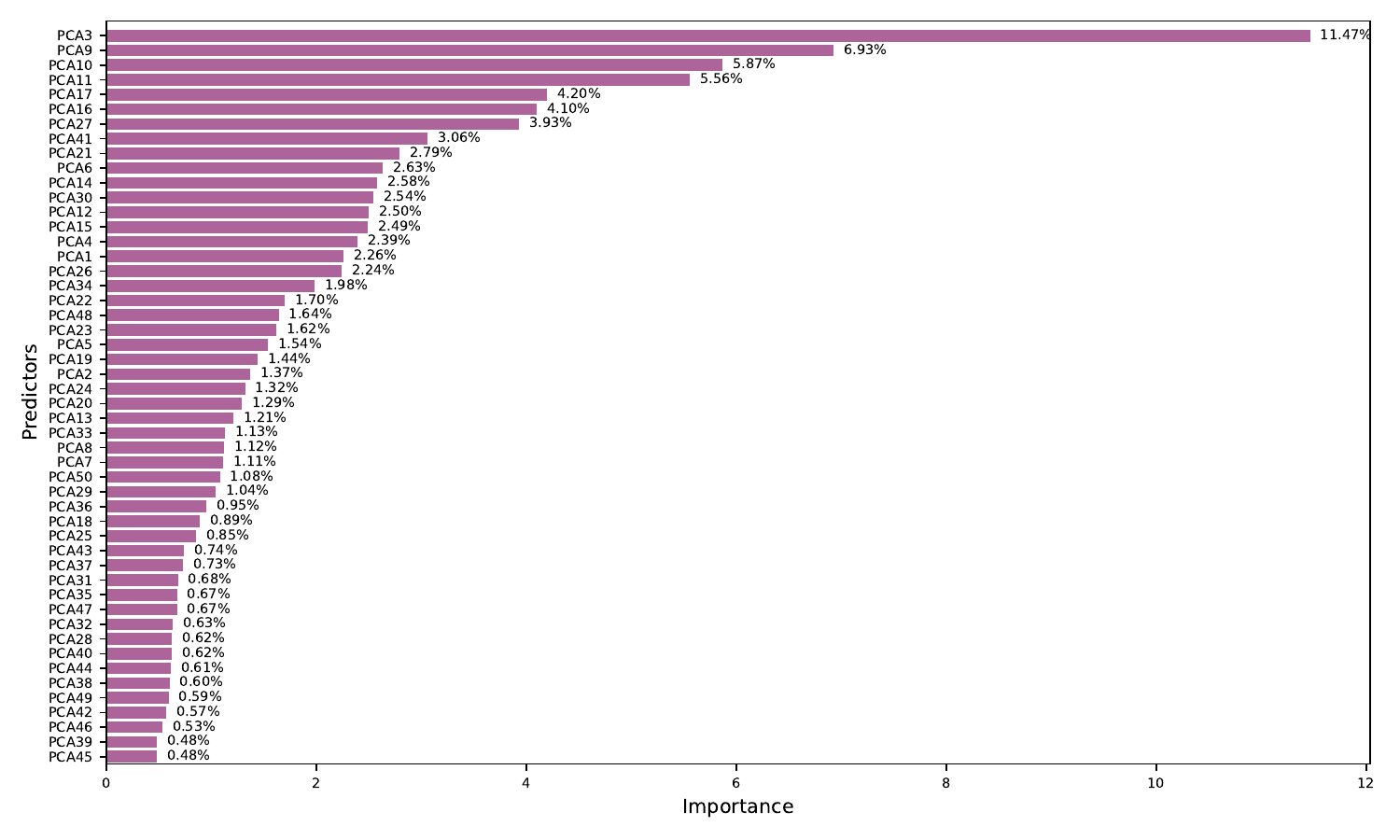}
\caption{Importance ranking of predictors based on their contribution to the SDH classification model.}
\label{SDH_featur_imp}
\end{figure}
\subsection{Feature Fusion and Feature Importance}
Based on a comprehensive review of prior studies, DL models with the highest frequency and best performance in Q1-ranked publications are selected as the most suitable candidates for feature extraction from CT scan images. In our study dataset, the CCA and VGG19 models demonstrate the best performance in feature extraction, while the cross-attention and EfficientNetB6 models show the worst performance in this regard. To refine the extracted features, we apply principal component analysis (PCA) and select the top 50 components from each model that account for the greatest data variance, considering these as informative features. This transformation enables a shift from basic features to more sophisticated ones, effectively reducing feature dimensionality and substantially lowering computational costs. To evaluate the discriminative power of these components, the Bootstrap Forest algorithm is employed. This algorithm leverages permutation techniques to determine the importance of each component in the final prediction by iteratively permuting the values of each component while keeping others constant. The greater the effect of a component’s permutation on model accuracy, the more critical that component is deemed to be. In fact, we define the Separation Index—often define as the ratio of intra-class to inter-class variance—as the component’s impact on model accuracy.

Fig. \ref{NICH_featur_imp} and Fig. \ref{SDH_featur_imp} illustrate the importance of each component of the CCA model, identified as the best-performing model, in detecting ICH and SDH, respectively. Components contributing less than $1\%$ to the overall performance were identified and excluded as non-discriminative. Consequently, 22 components were selected for ICH detection and 32 components for SDH detection, serving as the latent feature space to feed the boosting neural network.
\begin{table}[t]
\centering
\renewcommand{\arraystretch}{1.5}
\caption{Hyperparameters of Each Component Model in the Boosting Neural Network}
\label{hyperparameter_boosting_NN}
\resizebox{0.4\textwidth}{!}{ 
\begin{tabular}{|c|c|c|c|c|c|c|}
\hline
\multirow{2}{*}{Hidden} & \multicolumn{3}{c|}{Activation Function} & \multirow{2}{*}{Learning} & \multirow{2}{*}{Epochs} & \multirow{2}{*}{Penalty} \\ 
\cline{2-4} 
Layers & Sigmoid & Identity & Radial & Rate &  & Method \\
\hline
$L_1$ & 25 & 10 & 15 & \multirow{2}{*}{0.1} & \multirow{2}{*}{100} & \multirow{2}{*}{Squared} \\
$L_2$ & 10 & 5 & 10 &  &  & \\
\hline
\end{tabular}
}
\end{table}
\subsection{Boosting Neural Networks}
After identifying the informative and discriminative features through PCA and permutation techniques, these features are fed into a boosting neural network. This architecture comprises 10 component models, each focusing on correcting the errors of its predecessor. The final outputs of the component models are combined in a weighted manner based on accuracy, generating a probability for each slice belonging to each class, referred to as the original confidence map. Table \ref{hyperparameter_boosting_NN} details the hyperparameters of each component model. Our criterion for selecting 10 component models are based on starting with the minimum number of models and stopping when adding a new model component has a negligible effect on accuracy. Essentially, the number of component models represents a trade-off between complexity and performance.
\begin{figure}[t]
\centering
\includegraphics[scale=0.35]{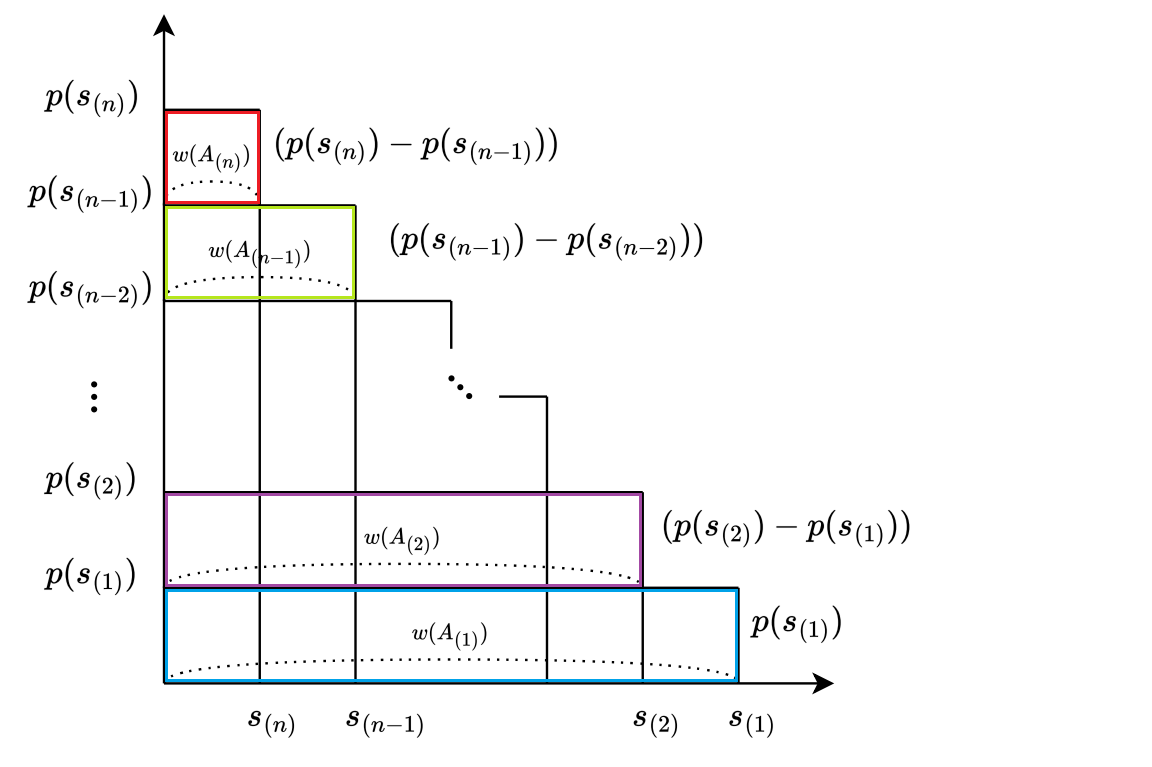}
\caption{Fuzzy integral operator framework for aggregation of diverse CT scan slices.}
\label{graph_choquet}
\end{figure}
\subsection{uncertainty-based fuzzy integral operator}
Since hemorrhage may not be visible in a single slice but could be apparent across consecutive slices, and single-slice diagnosis is highly susceptible to environmental variations such as changes in patient angle and position, analyzing multiple slices allows the AI system to examine a broader range of patterns, enhancing diagnostic accuracy. Therefore, to integrate multiple slices of a CT scan, we utilize an uncertainty-based fuzzy integral operator, which is capable of accounting for interactions among different slices and provides an adequate level of complexity \cite{klement2009universal,tahani1990information}.
\begin{flushleft}
    As illustrated in Fig. \ref{graph_choquet}, consider a set $S = \{s_1, s_2, s_3, \dots, s_n\}$ consisting of $n$ slices with probabilities $p(s_1), p(s_2), p(s_3), \dots, p(s_n)$, ordered such that:
    \begin{equation}
    p(s_1) \leq p(s_2) \leq \dots \leq p(s_n), \quad p(s_0) = 0
    \end{equation}
    The fuzzy measures are denoted as $w(A_1), w(A_2), \dots, w(A_n)$, and the parameter $\lambda$ is computed according to (\ref{lambda_eq}).
\end{flushleft}
\begin{equation} 
\lambda + 1 = \prod_{i=1}^{n}(1 + \lambda \cdot w_{\lambda}(s_i)), \quad -1 < \lambda < +\infty 
\label{lambda_eq} 
\end{equation}
If  $\sum_{i=1}^{n}w_{\lambda}(s_i) = 1$, $\lambda = 0$ is an acceptable value. When $\sum_{i=1}^{n}w_{\lambda}(s_i) <  1$,  $\lambda$  lies in the range $0 < \lambda < + \infty  $. Conversely, if  $\sum_{i=1}^{n}w_{\lambda}(s_i) >   1$, then $\lambda$ is in the range $-1 < \lambda < 0$. The weight $w(s_i)$ represents the assigned weight for the i-th slice.

The area of the first rectangle is $p(s_1).w(A_1)$, where:
\begin{multline}
w(A_1) = w(\{s_1, s_2, s_3, \dots, s_n\}) = w(s_1) + w(s_2)\\ + \dots + w(s_n) 
+ \lambda \cdot w(s_1) w(s_2) \dots w(s_n)
\end{multline}
\begin{flushleft}
The area of the second rectangle is $(p(s_2)-p(s_1)).w(A_2)$, where:
\begin{multline}
w(A_2) = w({s_2, s_3, \dots, s_n}) = w(s_2) + w(s_3)\\ + \dots + w(s_n) + \lambda \cdot w(s_2) w(s_3) \dots w(s_n) 
\end{multline}
Similarly, the area of the third rectangle is $(p(s_3)-p(s_2)).w(A_3)$, where:
\begin{multline} 
w(A_3) = w({s_3, s_4, \dots, s_n}) = w(s_3) + w(s_4)\\ + \dots + w(s_n) + \lambda \cdot w(s_3) w(s_4) \dots w(s_n) 
\end{multline}
Finally, the area of the n-th rectangle is $(p(s_n)-p(s_{n-1})).w(A_n)$, where:
\begin{equation} 
w(A_n) = w({s_n}) 
\end{equation}
The total aggregated area can be expressed as:
\begin{multline} 
\int p , dw = p(s_1) \cdot w(A_1) + (p(s_2) - p(s_1)) \cdot w(A_2)\\ + \dots + (p(s_n) - p(s_{n-1})) \cdot w(A_n) 
\end{multline}
This is generalized for all slices using (\ref{summation}):
\begin{equation}
\int p , dw = \sum_{i=1}^{n} (p(s_i) - p(s_{i-1})) \cdot w(A_i) \label{summation} \end{equation}
\end{flushleft}
The challenges associated with aggregating information from different CT slices using the standard fuzzy integral operator \cite{horanska2018generalization,ayub2009choquet,sugeno2014way} are as follows:

 \textbf{Computing the Parameter $\bm{\lambda}$:} According to (\ref{lambda_eq}), if a CT scan consists of $n$ slices, aggregating these slices using the fuzzy integral operator requires solving an n-degree  equation. Since $n$  may vary between different CT scans, $n \in \{c_1, c_2, ..., c_N\}$ where $c_k$  represents the number of slices in the k-th scan, each scan necessitates solving a unique equation of order $n$, leading to significant computational complexity.

Since $\sum_{i=1}^{n}w_{\lambda}(s_i) >   1$, to reduce the complexity, we treat $\lambda$ as a hyperparameter within the range $-1 < \lambda < 0$ and optimize it using a grid search approach. This method eliminates the need for solving high-order equations while maintaining computational efficiency.

\textbf{Assigning Weights to CT Scan Slices:} In the standard version of the fuzzy integral operator, the weight assigned to each information source (different slices of a CT image) is determined by a domain expert. To determine the significance of each slice, we use uncertainty as the guiding principle. In classification models, high confidence in a class prediction (e.g., one probability close to 1 and the others close to 0) indicates low uncertainty, while comparable probabilities for multiple classes suggest high uncertainty. Consequently, slices with lower uncertainty are assigned higher weights, and vice versa. For binary classification problems, the predicted probabilities for the two classes of the i-th slice of the j-th CT scan are denoted as $p_{i,j,1}$ and $p_{i,j,2}$. The weight for each slice is computed using (\ref{weight_binary}): 
\begin{equation}
w_{ij} = \max(p_{i,j,1}, p_{i,j,2}) - \min(p_{i,j,1}, p_{i,j,2}) \label{weight_binary} 
\end{equation}
In this approach, the fuzzy integral operator assigns higher weights to slices that provide more confident information about the occurrence or absence of hemorrhage.

After determining the probability of each slice belonging to each class, referred to as the confidence vector, using a boosting neural network, and assigning a weight to each slice as the belief value, the slices of the CT scan images are combined based on (\ref{summation}) to determine the final label of each scan. The proposed fuzzy integral operator, through the weighted aggregation of different CT image slices, improves hemorrhage detection accuracy while addressing the challenges associated with the standard version of the operator.
\section{Findings and Analysis}
\label{Results}
Since multiple CT scan images were typically recorded for each patient, to maximize the use of available scans and prevent data leakage, we included all scans from the same patient within the training data. The test data was designed to ensure subject independence. At the slice level, 80\% of the data was designated for training, and 20\% for testing. Considering that the training data included patients with multiple scans, this slice-level split corresponds to training the model on 50\% of the patients and testing it on the remaining 50\% at the patient level.
\subsection{ICH Detection}
The goal is to evaluate an intelligent model using CT scan images that can automatically detect the presence or absence of ICH. This is formulated as a binary classification problem, where the classes are defined as "Normal" and "ICH".
Tables \ref{table_nrml_ich_metricB} and \ref{table_nrml_ich_metricA} evaluate different models for slice-level ICH detection. Table \ref{table_nrml_ich_metricB} highlights regression metrics, where the CCA model achieved the best performance with the highest Generalized R-square (0.849) and lowest RASE (0.231) and MAD (0.118). In contrast, models like Cross Attention showed lower predictive capabilities.

Table \ref{table_nrml_ich_metricA} focuses on classification metrics, with the CCA model again leading, achieving the highest Accuracy (0.911), Precision (0.943), and F1-score (0.93). VGG16 also performed well with an Accuracy of 0.896 and F1-score of 0.915. These results confirm the superior performance of the CCA model for ICH detection.

\begin{table}[t]
    \centering
    \caption{Performance metrics of different models for slice-level ICH detection on the test set, including Generalized R-square, Entropy R-square, Root average squared error (RASE), Mean absolute deviation (MAD), and Log-Likelihood.}
    \label{table_nrml_ich_metricB}
    \resizebox{0.4\textwidth}{!}{
    \begin{tabular}{|l|c c c c c|}
    \hline
    \textbf{Model} & \multicolumn{5}{c|}{\textbf{Test Set}} \\
    \cline{2-6}
    & \textbf{Generalized R-square} & \textbf{Entropy R-square} & \textbf{RASE} & \textbf{MAD} & \textbf{Log-Likelihood} \\
    \hline
    VGG19       & 0.765  & 0.615  & 0.279 & 0.178 & -2806 \\
    ResNet50    & 0.759  & 0.607  & 0.283 & 0.187 & -2864 \\
    CCA   & 0.849  & 0.73   & 0.231 & 0.118 & -1964 \\
    Pyramid ViT     & 0.754  & 0.602  & 0.287 & 0.18  & -2905 \\
    DeiT        & 0.673  & 0.507  & 0.323 & 0.228 & -3592\\
    DenseNet       & 0.779  & 0.633  & 0.273 & 0.153 & -2675 \\
    EfficientNetB6      & 0.652  & 0.484  & 0.334 & 0.241 & -3761 \\
    Incep-ResNet     & 0.616  & 0.448  & 0.344 & 0.273 & -4029 \\
    Xception    & 0.755  & 0.602  & 0.286 & 0.166 & -2899 \\
    InceptionV3   & 0.767  & 0.618  & 0.278 & 0.169 & -2784 \\
    MobileNetV2      & 0.704  & 0.541  & 0.312 & 0.21  & -3345 \\
    Cross Attention   & 0.541  & 0.375  & 0.372 & 0.294 & -4557 \\
    ResNet152   & 0.753  & 0.6    & 0.288 & 0.181 & -2917 \\
    VGG16      & 0.819  & 0.687  & 0.251 & 0.138 & -2278 \\
    \hline
    \end{tabular}
    }
\end{table}

\begin{table}[t]
\centering
\caption{Evaluation metrics of different models for slice-level ICH detection, including Accuracy, Precision, Sensitivity, Specificity, and F1-score.}
\label{table_nrml_ich_metricA}
\resizebox{0.4\textwidth}{!}{
\begin{tabular}{|c|c|c|c|c|c|}
\hline
\textbf{Model} & \textbf{Accuracy} & \textbf{Precision} & \textbf{Sensitivity} & \textbf{Specificity} & \textbf{F1-score} \\ \hline
VGG19     & 0.878    & 0.911     & 0.883       & 0.893       & 0.897    \\ 
ResNet50  & 0.874    & 0.912     & 0.872       & 0.895       & 0.891    \\ 
CCA  & 0.911    & 0.943     & 0.917       & 0.925       & 0.93    \\ 
Pyramid ViT   & 0.868    & 0.903     & 0.869       & 0.887       & 0.886    \\ 
DeiT     & 0.84    & 0.874     & 0.84       & 0.859       & 0.857    \\ 
DenseNet121     & 0.88    & 0.919     & 0.878       & 0.902       & 0.898    \\ 
EfficientNetB6   & 0.824    & 0.859     & 0.823       & 0.845       & 0.841   \\ 
Incep-ResNet   & 0.821    & 0.872     & 0.8       & 0.862       & 0.835    \\ 
Xception  & 0.873    & 0.916     & 0.865       & 0.901       & 0.89    \\ 
InceptionV3 & 0.879    & 0.912     & 0.882       & 0.895       & 0.897    \\ 
MobileNetV2    & 0.846    & 0.882     & 0.846       & 0.866       & 0.864    \\ 
Cross Attention & 0.781    & 0.816     & 0.776       & 0.805       & 0.795    \\ 
ResNet152 & 0.867    & 0.901     & 0.87       & 0.884       & 0.885    \\ 
VGG16       & 0.896    & 0.924     & 0.907       & 0.905       & 0.915  \\ \hline
\end{tabular}
} 
\end{table}

\begin{figure}[t]
\centering
\includegraphics[scale=0.35]{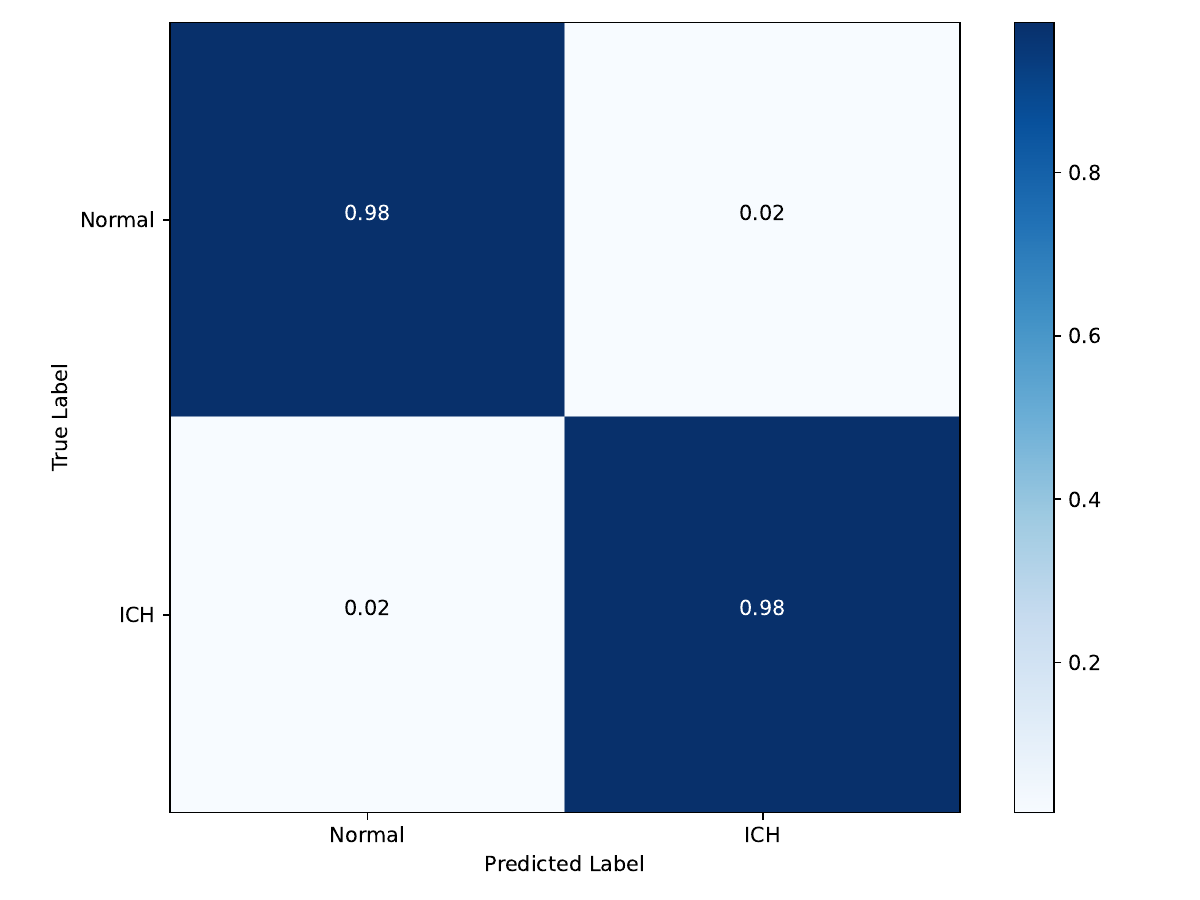}
\caption{Scan-level confusion matrix on test data for ICH detection.}
\label{NICH_confusion_matrix}
\end{figure}

Fig. \ref{NICH_confusion_matrix} illustrates the confusion matrix of the CCA model at the scan-level, where different slices were combined using the fuzzy integral operator based on uncertainty in the detection of ICH. With an accuracy of 98.4\%, sensitivity of 98.4\%, precision of 98\%, specificity of 98.4\%, and F1-score of 98.15\%, the proposed approach demonstrates strong performance characteristics.

\subsection{SDH Detection}
The aim here is to accurately identify the specific type of SDH. This binary classification includes "ASDH" and "CSDH" classes.

Table \ref{table_sdh_type_metricB} focuses on regression-based metrics. The CCA model demonstrates the best performance with the highest Generalized R-square (0.85) and the lowest MAD (0.087). It significantly outperforms models like EfficientNetB6, which has the lowest Generalized R-square (0.629) and highest error metrics.

Table \ref{table_sdh_type_metricA} highlights classification metrics. The CCA model again leads with the highest Accuracy (0.913). The Incep-ResNet and Pyramid ViT models also perform well, while models like EfficientNetB6 and Xception show relatively lower performance.

In summary, the CCA model consistently achieves superior results across both regression and classification metrics, making it the most effective for slice-level ICH and SDH detection.

\begin{table}[t]
    \centering
    \caption{Performance metrics of different models for slice-level SDH detection on the test set, including Generalized R-square, Entropy R-square, RASE, MAD, and Log-Likelihood.}
    \label{table_sdh_type_metricB}
    \resizebox{0.4\textwidth}{!}{
    \begin{tabular}{|l|c c c c c|}
    \hline
    \textbf{Model} & \multicolumn{5}{c|}{\textbf{Test Set}} \\
    \cline{2-6}
    & \textbf{Generalized R-square} & \textbf{Entropy R-square} & \textbf{RASE} & \textbf{MAD} & \textbf{Log-Likelihood} \\
    \hline
    VGG19       & 0.778 & 0.634 & 0.276 & 0.156 & -233 \\
    ResNet50    & 0.808 & 0.674 & 0.26 & 0.152 & -207 \\
    CCA   & 0.85 & 0.734 & 0.722 & 0.087 & -169 \\
    Pyramid ViT     & 0.813 & 0.691 & 0.245 & 0.119 & -82 \\
    DeiT        & 0.765 & 0.627 & 0.272 & 0.163 & -98 \\
    DenseNet121       & 0.807 & 0.672 & 0.257 & 0.156 & -208 \\
    EfficientNetB6      & 0.629 & 0.462 & 0.34 & 0.235 & -342 \\
    Incep-ResNet     & 0.841 & 0.72 & 0.233 & 0.126 & -178 \\
    Xception    & 0.68 & 0.517 & 0.323 & 0.213 & -184 \\
    InceptionV3   & 0.728 & 0.571 & 0.298 & 0.171 & -273 \\
    MobileNetV2      & 0.74 & 0.585 & 0.295 & 0.164 & -264 \\
    Cross Attention   & 0.683 & 0.53 & 0.303 & 0.197 & -124 \\
    ResNet152   & 0.817 & 0.686 & 0.248 & 0.121 & -199 \\
    VGG16     & 0.801 & 0.675 & 0.25 & 0.143 & -86   \\
    \hline
    \end{tabular}
    }
\end{table}

\begin{table}[t]
    \centering
    \caption{Evaluation metrics of different models for slice-level SDH detection, including Accuracy, Precision, Sensitivity, Specificity, and F1-score.}
    \label{table_sdh_type_metricA}
    \resizebox{0.4\textwidth}{!}{
    \begin{tabular}{|c|c|c|c|c|c|}
    \hline
    \textbf{Model} & \textbf{Accuracy} & \textbf{Precision} & \textbf{Sensitivity} & \textbf{Specificity} & \textbf{F1-score} \\ \hline
VGG19     & 0.867    & 0.892     & 0.907       & 0.842       & 0.899    \\ 
ResNet50  & 0.875    & 0.889     & 0.926       & 0.834       & 0.907    \\ 
CCA  & 0.913    & 0.94     & 0.94       & 0.905       & 0.94    \\ 
PVT   & 0.903    & 0.926     & 0.9       & 0.88      & 0.912    \\ 
Deit      & 0.879    & 0.893     & 0.878       & 0.921       & 0.885    \\ 
DenseNet121     & 0.887    & 0.89     & 0.952       & 0.832       & 0.92    \\ 
EfficientNetB6    & 0.815    & 0.845     & 0.863       & 0.781       & 0.854    \\ 
Incep-ResNet   & 0.91    & 0.941     & 0.932       & 0.907       & 0.937    \\ 
Xception  & 0.826    & 0.843     & 0.844       & 0.824       & 0.844    \\ 
InceptionV3 & 0.861    & 0.882     & 0.907       & 0.827       & 0.894    \\ 
MobileNetV2    & 0.855    & 0.878     & 0.901       & 0.822       & 0.889    \\ 
Cross Attention & 0.857    & 0.865     & 0.859       & 0.839       & 0.862    \\ 
ResNet152 & 0.881    & 0.948     & 0.909       & 0.917       & 0.928    \\ 
VGG16   & 0.898    & 0.911     & 0.905       & 0.885       & 0.908     \\ \hline
    \end{tabular}
    }    
\end{table}

\begin{figure}[t]
\centering
\includegraphics[scale=0.35]{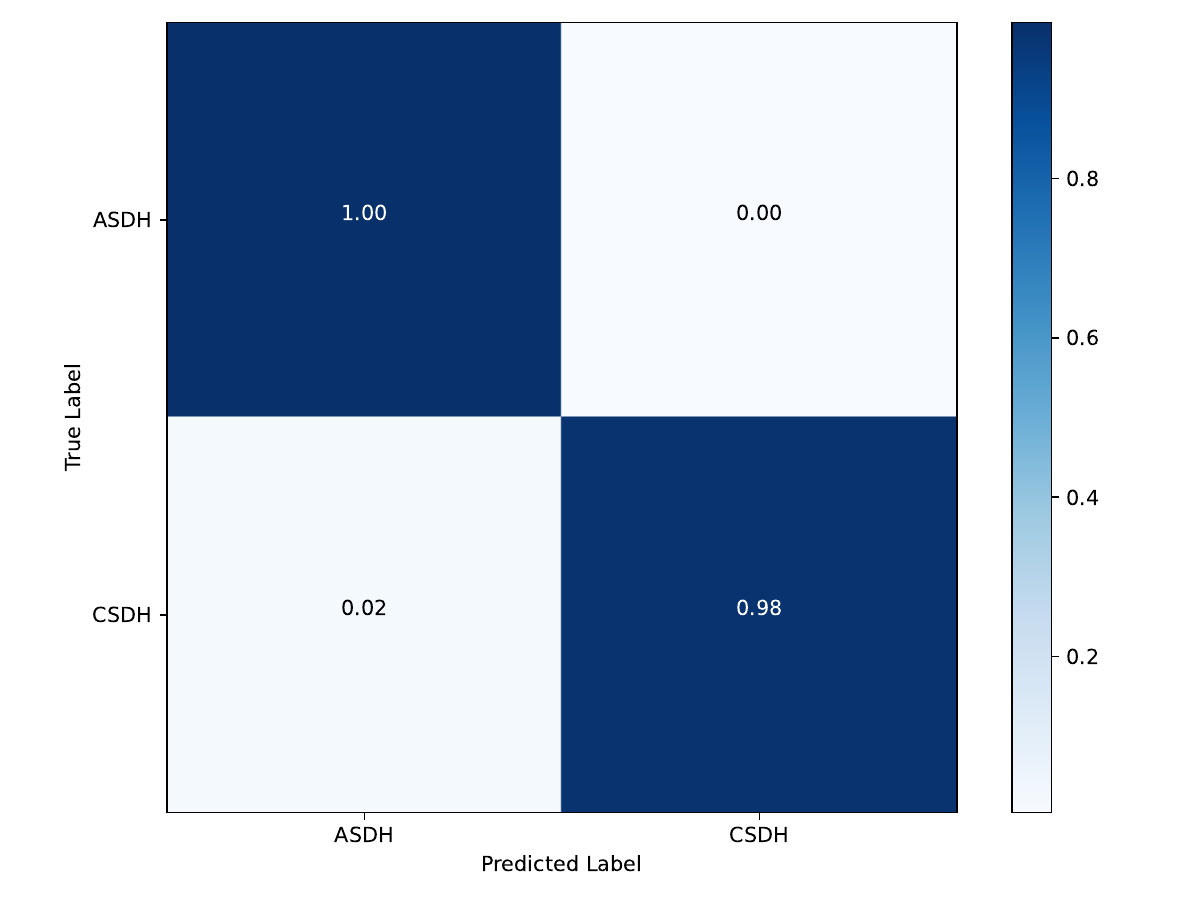}
\caption{Scan-level confusion matrix on test data for SDH detection.}
\label{SDH_confusion_matrix}
\end{figure}

Fig. \ref{SDH_confusion_matrix} depicts the confusion matrix of the CCA model at the scan-level, in which diverse slices are integrated using the proposed fuzzy integral operator to detect SDH. The accuracy of 98.8\%, sensitivity of 98.7\%, precision of 98.6\%, specificity of 98.7\%, and F1-score of 98.65\% indicate that the classification model performs exceptionally well, with minimal misclassification, achieving near-perfect accuracy in both class predictions.

Table \ref{disscusion_table} provides a comparison of this study with other research employing deep neural networks for ICH analysis. Although prior studies have achieved commendable results, our approach demonstrates that integrating PCA and feature importance techniques not only reduces model complexity but also enhances accuracy. Notably, this also significantly improves the model's explainability.

\begin{table}[t]
\caption{Comparison summary of the proposed method with existing approaches.}
\label{disscusion_table}
\centering
\resizebox{0.4\textwidth}{!}{
\begin{tabular}{|l|l|l|l|l|l|l|l|}
\hline
Work &
  Year & Dataset &
  \begin{tabular}[c]{@{}l@{}}Number of \\ Samples\end{tabular} &
  Objective &
  Deep Learning Model &
  \begin{tabular}[c]{@{}l@{}}Validation\\  Strategy\end{tabular} &
  Performance \\ \hline
{\cite{kyung2022improved}} &
  2022 & Public &
  1469 scans &
  Normal/ICH &
  \begin{tabular}[c]{@{}l@{}}Pre-trained encoder-\\  LSTM\end{tabular} &
  Single-fold &
  \begin{tabular}[c]{@{}l@{}}AUC= 0.988,\\  F1-score = 0.933\end{tabular} \\ \hline
{\cite{lopez2022deep}} &
  2022 & Public &
  1540 scans &
  Normal/ICH &
  CNN with attention &
  Single-fold &
  \begin{tabular}[c]{@{}l@{}}F1-score = 0.839,\\   AUC  = 0.957\end{tabular} \\ \hline
{\cite{gudadhe2023classification}} &
  2023 & Public &
  2501 slices &
  Normal/ICH &
  \begin{tabular}[c]{@{}l@{}}Weber Local Descriptor-\\ Random Forest\end{tabular} &
  Single-fold &
  \begin{tabular}[c]{@{}l@{}}Acc= 0.86,\\  Sens= 0.87\end{tabular} \\ \hline

  {\cite{cortes2023deep}} &
  2023 & Public &
  18938 scans &
  Normal/ICH &
  \begin{tabular}[c]{@{}l@{}}EfficientDet’s deep-learning \\technology  \end{tabular} &
  Single-fold &
  \begin{tabular}[c]{@{}l@{}}Acc= 0.964,\\ AUC = 0.978\end{tabular} \\ \hline
{\cite{malik2023computational}} &
  2023 & Public &
  491 patients &
  Normal/ICH &
  Deep fuzzy Network &
  5-fold &
  \begin{tabular}[c]{@{}l@{}}Acc = 0.94, Pre =0.938 \\ Sens = 0.926, F1-score = 0.88 \end{tabular} \\ \hline
{\cite{perez2024end}} &
  2024 & Public &
  1150 scans &
  Normal/ICH &
  \begin{tabular}[c]{@{}l@{}}An attention-based model\\ with gaussian processes\end{tabular} &
  Single-fold &
  \begin{tabular}[c]{@{}l@{}}Acc= 0.87, \\ F1-score = 0.88,\\  Prec= 0.82, \\ AUC = 0.96\end{tabular} \\ \hline
{\cite{lin2024semi}} &
  2024 & Public &
  457 scans &
  Normal/ICH &
  \begin{tabular}[c]{@{}l@{}}Patch FCN \\ with dilated ResNet38\end{tabular} &
  5-fold &
  AUC = 0.939 \\ \hline
 {\cite{shah2024human}} &
  2024 & Private &
  642 patients &
  Normal/ICH &
  \begin{tabular}[c]{@{}l@{}}ML powered Near-infrared \\ spectroscopy based (mNIRS)\\ system \\ \end{tabular}&
  Single-fold &
   \begin{tabular}[c]{@{}l@{}}Acc= 0.942, Sens = 0.948\\  Spec = 0.941 \end{tabular} \\ \hline
 {\cite{castro2024hyperbolic}} &
  2024 & Public &
  70000 slices &
  Normal/ICH &
  \begin{tabular}[c]{@{}l@{}}Multiple instance\\ learning with \\ Gaussian Processes \\ \end{tabular}&
  Single-fold &
   \begin{tabular}[c]{@{}l@{}}Acc= 0.915, AUC = 0.94\\  F1-score = 0.902 \end{tabular} \\ \hline
\begin{tabular}[c]{@{}l@{}}Our \\ work\end{tabular}  &
  2024 & Private &
  \begin{tabular}[c]{@{}l@{}} \\ 31596 slicess \\ (normal/ICH) \\ \\ \hline \\ 2785 slices \\ (SDH types) \\ \\ \end{tabular} &
  \begin{tabular}[c]{@{}l@{}}\\Normal/ICH \\ \\ \hline  \\  ASDH/CSDH \\ \\ \end{tabular} &
  \begin{tabular}[c]{@{}l@{}}Co-scale convolutional\\ attention model \\ with uncertainty-based fuzzy \\ integral operator \\ \end{tabular} &
  Single-fold &
  \begin{tabular}[c]{@{}l@{}}\\ Acc=0.984, Prec=0.98 \\ Sens=0.984, Spec=0.984 \\ F1-score = 0.981 \\ \\ \hline \\ Acc=0.988, Prec=0.986, \\ Sens=0.987, Spec=0.987, \\ F1-score =0.986 \\ \\ \end{tabular} \\ \hline
\end{tabular}%
}
\end{table}
\section{Future Research Directions}
\label{recomm}
In the future, we aim to explore:
\begin{itemize}
    \item  An uncertainty-based fuzzy integral operator was utilized to fuse different slices of CT scan images at the decision level. Future studies can employ this operator at the data or feature level, allowing features extracted from various deep networks to be combined using this operator.
    \item   Future research could investigate the optimal number of principle components, explore more sophisticated dimensionality reduction approaches, and optimize other hyperparameters.
    \item We excluded CT scan images that showed the occurrence of more than one type of hemorrhage. Future studies may define a new class for such cases.
    \item It is recommended to use clinical data such as blood pressure, age, and injury area to develop a rapid classification task aimed at identifying likely cases of hemorrhage.
    \item Future studies can focus on implementing the proposed system in a clinical setting to showcase its real-world potential and limitations.
\end{itemize}
\section{Conclusion}
\label{Conclusion}
ICH requires prompt and accurate diagnosis to ensure timely implementation of appropriate treatments. In this study, after feature extraction from CT scan images, the most informative and most discriminative features were identified to feed into a hybrid neural network, enabling the classification of each slice into specific classes. Subsequently, we introduced a novel version of a fuzzy integral operator, enhanced with uncertainty-based weighting, to aggregate slices from CT scan images. The proposed approach, implemented on two separate datasets, demonstrated significant improvement in hemorrhage detection at the scan-level through the intelligent combination of multiple slices.

\bibliography{references}
\end{document}